\newif\ifshort
\shorttrue

\newif\ifacm
\acmfalse

\newif\ifanonymous
\anonymousfalse

\newif\ifdraft
\draftfalse

\newif\ifapponly
\apponlyfalse

\ifanonymous
\documentclass[sigconf,anonymous]{acmart}
\else
\documentclass[sigconf]{acmart}
\fi

\ifacm
\else
\setcopyright{none}
\renewcommand\footnotetextcopyrightpermission[1]{}
\fi

\AtBeginDocument{%
  }

\usepackage{listings}
\usepackage{xcolor}
\usepackage{boxedminipage}
\usepackage[l3]{csvsimple}
\usepackage{booktabs}
\usepackage{enumitem}

\lstdefinelanguage{json}{
    basicstyle=\ttfamily,
    numbers=left,
    numberstyle=\small,
    stepnumber=1,
    numbersep=8pt,
    showstringspaces=false,
    breaklines=true,
    frame=single,
    string=[s]{"}{"},
    stringstyle=\color{blue},
    comment=[l]{//},
    commentstyle=\color{gray},
    morecomment=[s]{/*}{*/},
    literate=
     *{0}{{{\color{magenta}0}}}{1}
      {1}{{{\color{magenta}1}}}{1}
      {2}{{{\color{magenta}2}}}{1}
      {3}{{{\color{magenta}3}}}{1}
      {4}{{{\color{magenta}4}}}{1}
      {5}{{{\color{magenta}5}}}{1}
      {6}{{{\color{magenta}6}}}{1}
      {7}{{{\color{magenta}7}}}{1}
      {8}{{{\color{magenta}8}}}{1}
      {9}{{{\color{magenta}9}}}{1}
      {:}{{{\color{black}{:}}}}{1}
      {,}{{{\color{black}{,}}}}{1}
      {\{}{{{\color{black}{\{}}}}{1}
      {\}}{{{\color{black}{\}}}}}{1}
      {[}{{{\color{black}{[}}}}{1}
      {]}{{{\color{black}{]}}}}{1},
}
  
\usepackage{comment}

\newcommand{\fsl}{\ifanonymous\textsf{XYZ}\else\textsf{FSL}\fi}
\newcommand{\sle}{\textsf{SLE}}
\newcommand{\swebok}{\textsf{SWEBOK}}

\newcommand{\myparagraph}[1]{\paragraph{\textbf{#1}}}
\newcommand{\dbfootnote}{\footnote{Anonymized for double-blind review.}}

\begin{document}

\title{Continuous Ontology Assurance through External Resources}
\subtitle{An Ontology Engineering Method with a Case Study in Software Language Engineering}

\author{Ralf L{\"a}mmel}
\affiliation{%
  \institution{University of Koblenz, Faculty of Computer Science}
  \city{Koblenz}
  \country{Germany}}

\renewcommand{\shortauthors}{L{\"a}mmel}

\newcommand{\coater}{`\emph{Continuous Ontology Assurance Through External Resources}'}
\newcommand{\mabbr}{\textsf{COATER}}

\begin{abstract}
  Domain ontologies require review against knowledge recorded in
  evolving external resources. We propose \coater{} (\mabbr), a method
  that manages executable review components called \emph{arrows}. Each
  arrow compares a scoped part of an ontology with a relevant resource
  and reports support or challenges to its axioms. Challenges may
  prompt proposals for completion, correction, or restructuring and
  become tasks for human or agent-assisted triage. Per-axiom coverage
  summarizes the assessments. We illustrate \mabbr{} with the \fsl{}
  ontology in software language engineering (\sle): arrows review
  language coverage, language conceptualization, and
  software-engineering activities. An LLM-assisted linkage exercise
  increases compliance with a defined Wikipedia-linkage policy from
  405 to 600 of 600 entities. A structured catalogue of \sle{}
  resources organizes review opportunities.
\end{abstract}

\maketitle

\ifacm
\else
\pagestyle{empty}
\fi

\section{Introduction}

In this paper, we describe an ontology engineering method for \coater{}
(\mabbr) and apply it to software language engineering (\sle) and a
corresponding domain ontology \fsl.\ifanonymous\dbfootnote{}\fi
Maintaining such an ontology requires more than checking its internal
structure: its assertions and coverage also need to be assessed
against knowledge of the domain. Ontology evaluation addresses
structural and functional quality~\cite{gangemi2006modelling};
test-driven Linked Data evaluation provides executable quality checks
and measures their coverage~\cite{kontokostas2014test}. External
resources can provide evidence for a different question: which parts
of a maintained ontology are supported, challenged, or missing
relative to a particular source?

\begin{figure}[htbp]
    \vspace{-37\in}
    \centering
    \includegraphics[width=0.8\columnwidth]{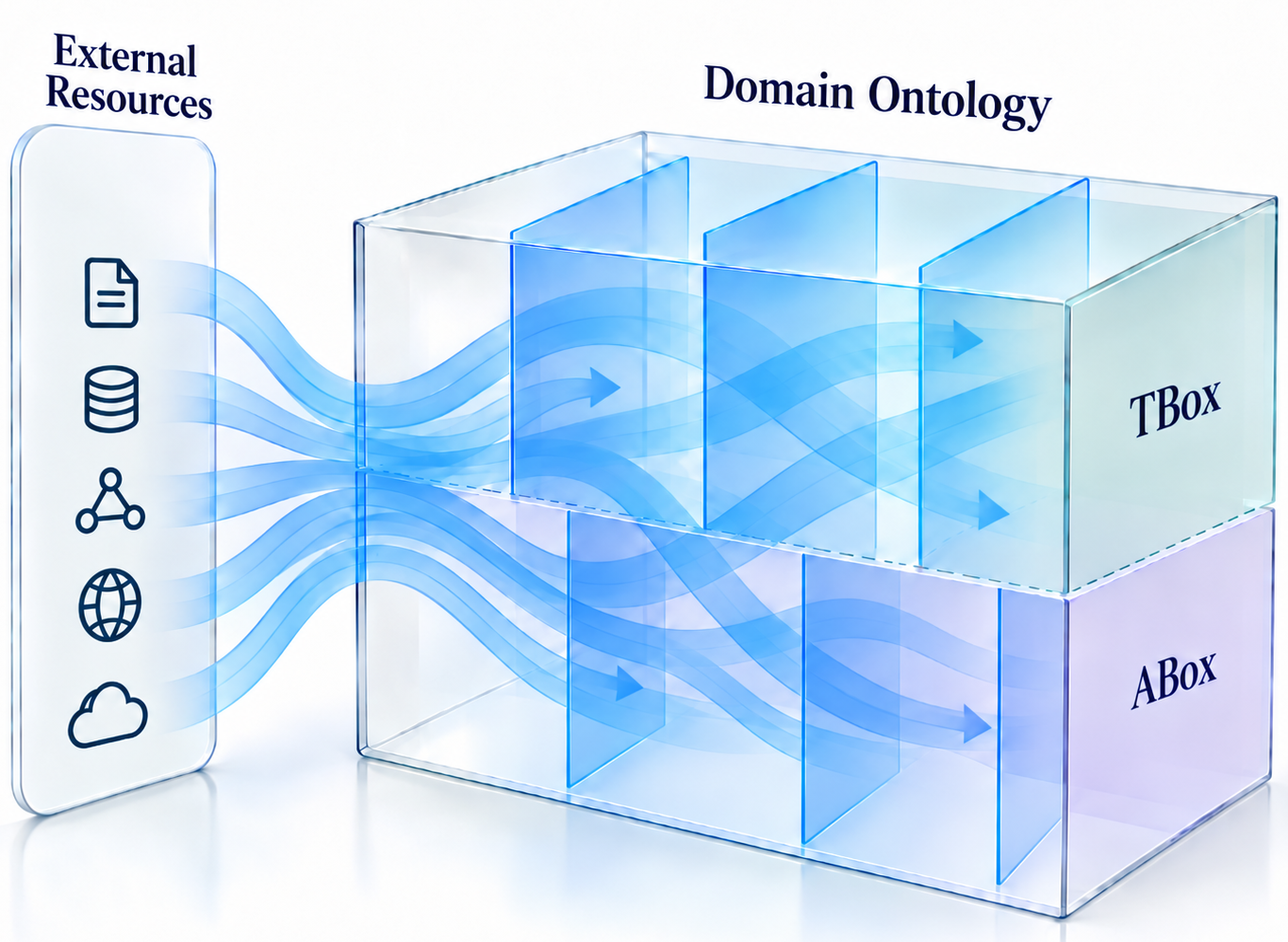}
    \caption{Arrows in the \mabbr{} method.}
    \label{fig:fslcube}
    \vspace{-37\in}
\end{figure}

Figure~\ref{fig:fslcube} illustrates the \mabbr{} method. On the right
is the domain ontology to be assured, with TBox and ABox parts
corresponding to the scope of external knowledge resources. On the
left are such resources, including ontologies, Wikipedia pages, Bodies
of Knowledge, and scholarly papers. The \emph{arrows} represent
ontology review components that use external resources to support or
challenge ontology elements. Challenges may include revision proposals
for completion, correction, or restructuring, and may give rise to
tasks handled through agentic triaging with a human in the loop. The
\mabbr{} method administers the continuous aggregation, execution, and
maintenance of such arrows for ontology assurance.

\vspace{-22\in}

\paragraph*{Contributions} Prior work uses external sources for
ontology evolution~\cite{zablith2009evolva} and studies LLM-assisted
ontology evaluation~\cite{lippolis2025assisting}. We contribute a
method for recurring, resource-scoped reviews of a maintained
ontology: arrows support or challenge its content and may propose
revisions for review. We define the arrow and assessment model,
including per-axiom coverage (Sec.~\ref{sec:meth}). In the case study,
we demonstrate ABox and TBox reviews, improved Wikipedia linkage, and
a catalogue of \sle{} resources for \fsl{}
(Sec.~\ref{sec:arrows}, Sec.~\ref{sec:improve}, Sec.~\ref{sec:cata}).

\vspace{-22\in}

\paragraph*{Road-map of the paper} Section~\ref{sec:arrows} presents
case-study arrows, Sec.~\ref{sec:meth} the general method,
Sec.~\ref{sec:improve} an assurance improvement, and
Sec.~\ref{sec:cata} a catalog of external knowledge resources.
Section~\ref{sec:related} discusses related work, and
Sec.~\ref{sec:concl} concludes.

{\sloppypar

\textbf{There is a \ifanonymous{}double-blind reviewing-compliant\fi{} reproducibility package
  available for this
  paper.\footnote{\url{https://figshare.com/s/71d24b1526965387d7e4}}
  Upon acceptance of this submission, an extended technical report
  will be published with appropriate linkage for the GitHub-based
  hosting of the \fsl{} ontology and the \mabbr{} case study.}

}


\section{Illustrative \mabbr{} Arrows}
\label{sec:arrows}

In this section, we begin with the demonstration of the \mabbr{}
method in the \sle{} domain for the \fsl{} ontology. We focus here on
the design and implementation of a few \mabbr{} arrows.
\ifanonymous\else{}These illustrative arrows can be found in the \fsl repository.\footnote{\url{https://github.com/softlang/fsl/tree/main/ekrs}}\fi


\subsection{A `Language Popularity' Arrow}
\label{sec:arrow:tiobe}

We start with a very simple arrow, where we aim to establish coverage
for popular programmnig languages by the ontology. We pick the
TIOBE Programming Community
Index\footnote{\url{https://www.tiobe.com/tiobe-index/}} as an
established resource for ranking programming languages.

\myparagraph{Resource Extraction}

We retrieve TIOBE's top-10 languages with a Python-based scraper like this:

\begin{lstlisting}[language=Python,basicstyle=\ttfamily\scriptsize,numbers=none]
tiobe = pd.read_html("https://www.tiobe.com/tiobe-index/")
tiobe = tiobe[0].head(10)
tiobe = tiobe.iloc[:, [0, 1, 4]]
tiobe.columns = ["Rank", "RankLastYear", "Language"]
\end{lstlisting}

These are the top 10 at the time of writing:

\begin{center}\scriptsize
\begin{tabular}{|r|r|l|}
\hline
Rank & Rank Last Year & Language \\
\hline
\begin{filecontents*}{snippets/top10.csv}
Rank,RankLastYear,Language
1,1,Python
2,3,C
3,2,C++
4,4,Java
5,5,C#
6,6,JavaScript
7,7,Visual Basic
8,11,SQL
9,13,R
10,18,Rust
\end{filecontents*}
\csvreader[
  head to column names,
  respect sharp,
  late after line=\\
]{snippets/top10.csv}{}{%
  \Rank & \RankLastYear & \detokenize\expandafter{\Language}
}%
\hline
\end{tabular}
\end{center}

\myparagraph{Relevant Ontology Scope}

We query the ontology like this:

\begin{lstlisting}[language=Sparql,basicstyle=\ttfamily\scriptsize,numbers=none]
SELECT DISTINCT ?language ?classifier
WHERE {
    ?classifier rdfs:subClassOf* tbox:SoftwareLanguage .
    ?language rdf:type ?classifier .
}
ORDER BY ?classifier ?language
\end{lstlisting}

In this query, we include instances of all subclasses of
`SoftwareLanguages'~---~including `ProgrammingLanguage', as we
noticed that Tiobe views SQL as a programming language, while \fsl{}
assumes a nuanced taxonomy of software languages.

\myparagraph{Ontology Review} Let's test for completeness:

\begin{lstlisting}[language=Python,basicstyle=\ttfamily\scriptsize,numbers=none]
onto = ... # Query ontologies as shown
missing = tiobe[~tiobe["Language"].isin(onto["SoftwareLanguage"])]
\end{lstlisting}

We get these results:

\begin{center}\scriptsize
\begin{tabular}{r|r|l}
\hline
Rank & Rank Last Year & Language \\
\hline
\begin{filecontents*}{snippets/top10_missing_initial.csv}
Rank,RankLastYear,Language
3,2,C++
5,5,C#
7,7,Visual Basic
\end{filecontents*}
\csvreader[
  head to column names,
  respect sharp,
  late after line=\\
]{snippets/top10_missing_initial.csv}{}{%
  \Rank & \RankLastYear & \detokenize\expandafter{\Language}
}%
\hline
\end{tabular}
\end{center}

These are false negatives, i.e., these languages are not missing in
the ontology; they are merely showing up due to naming/lexical
conventions. We reconcile the names:

\begin{lstlisting}[language=Python,basicstyle=\ttfamily\scriptsize,numbers=none]
tiobe["Language"] = tiobe["Language"].replace({
    "C#": "C_Sharp",
    "C++": "Cpp",
    "Visual Basic": "Visual_Basic"})
\end{lstlisting}

With these renamings in place, the ontology misses none of the
languages retrieved from the external resource. In the terminology of
the \mabbr{} method, 10 languages from the ontology are `supported'
by TIOBE.

\myparagraph{Discussion}

This `hello world' example is very limited. Trivially, we looked only
at the top-10 languages on TIOBE for the sake of presentation. Let's
identify issues to be addressed eventually for a comprehensive
discussion of the \mabbr{} method.

\begin{description}[leftmargin=1em]

\item[ABox only] The TBox was not involved here other than the
  realization that knowledge resources may disagree on the extension
  of the concept `Programming language' which in turn led us to target
  software languages instead of programming languages in \fsl.
  
\item[Completeness only] We only assess the ontology’s completeness
  relative to the external knowledge resource. We do not verify
  whether its language-related assertions are supported by TIOBE; this is
  addressed in the next arrow demonstration
  (Sec.~\ref{sec:pediaarrow}).

\item[Threats to validity] We bet on name equality and similarity; we
  also rely on expert insight on the assumed correspondence of
  language categories.
 
\end{description}


\subsection{A `Language Conceptualization' Arrow}
\label{sec:pediaarrow}

A more advanced assurance scenario follows. \fsl{} classifies and
conceptualises software languages. In the case of programming
languages, this includes familiar notions such as programming
paradigms and typing disciplines. The corresponding conceptualization
relies on nuanced properties and extra classes for concepts. We
leverage now an LLM to consult Wikipedia as an external knowledge
resource for ontology review.


\begin{figure}[htbp]\scriptsize
\begin{boxedminipage}{\linewidth}
\textbf{Relevant Ontology Scope}
  
\begin{boxedminipage}{\linewidth}\itshape
  We are concerned with the entity \{language\} of the ontology
  \{ontology\}. We focus as follows. In particular, see the tbox
  here: \{tbox\}. Also, see the the part of the abox where the entity
  \{language\} appears as subject or object; see this CSV file:
  \{abox\}.
\end{boxedminipage}

\smallskip

\textbf{Knowledge Extraction}

\begin{boxedminipage}{\linewidth}\itshape
  We are also concerned with the Wikipedia page for the entity
  \{language\}, i.e.: \{wikipedia\}. Let's focus on all extractable
  statements that may be generally of interest for a domain ontology
  like \{ontology\}, without yet commiting to the particular TBox at
  hand.
\end{boxedminipage}

\smallskip

\textbf{Ontology Review}

\begin{boxedminipage}{\linewidth}\itshape
Let's compare the ontology with the Wikipedia page. The result should be an array of JSON objects with key-values pairs as follows:
\begin{itemize}
\item key "subject", value "\{language\}" or another entity that is or could be in the ontology.
\item key "property", value the name of a property declared in the tbox or exercised in the abox fragment at hand.
\item key "object", value "\{language\}" or another entity that is or could be in the ontology.
\item key "assessment", value "support", "challenge", "reject", "add"
  to mean that an existing triple from the given context is supported
  by Wikipedia, must be rejected or challenged on the grounds of
  Wikipedia, or should be added based on Wikipedia, in alignment with
  the tbox at hand.
\item key "explanation", value a short paragraph explaining the
  assessment, possibly quoting Wikipedia content as means of support.
\end{itemize}
\end{boxedminipage}
\end{boxedminipage}
\vspace{-37\in}
\caption{Prompt for Wikipedia-based ontology review.}
\label{fig:pedia-prompt}
\vspace{-37\in}
\end{figure}


\begin{figure}[htbp]\scriptsize

\bigskip

\hfill{}Support
\vspace{-99\in}
\begin{lstlisting}[language=json,basicstyle=\ttfamily\scriptsize,numbers=none]
{
"subject": "Java",
"property": "nativelySupports",
"object": "Exceptions",
"assessment": "support",
"explanation": "Wikipedia explicitly discusses Java exceptions, including that calling methods for a non-existent object can cause a null pointer exception, and describes Java packages as containing exceptions. This directly supports exceptions as a native Java language capability."
}
\end{lstlisting}

\medskip

\hfill{}Challenge
\vspace{-99\in}
\begin{lstlisting}[language=json,basicstyle=\ttfamily\scriptsize,numbers=none]
{
"subject": "Java",
"property": "nativelySupports",
"object": "AdHocPolymorphism",
"assessment": "challenge",
"explanation": "Java has forms of polymorphism associated with methods and object-oriented programming, but the Wikipedia page explicitly notes that Java does not support operator overloading. Because AdHocPolymorphism can cover several mechanisms and the page does not directly characterize Java in those terms, native support should be treated cautiously rather than rejected."}
\end{lstlisting}
\vspace{-42\in}
\caption{LLM Response for Wikipedia-Ontology Alignment}
\label{fig:pedia-response}
\vspace{-37\in}
\end{figure}


\myparagraph{End-to-end Prompting}

Figure~\ref{fig:pedia-prompt} illustrates the ontology-review prompt,
structured around ontology scope, knowledge extraction, and review. The
LLM considers all facts about a language in the ontology, extracts
conceivable facts from Wikipedia, and assesses each ontology fact.
Figure~\ref{fig:pedia-response} shows one support and one challenge.

\myparagraph{Assessments as Tasks}

Support for `exceptions' is plausible. Challenging `ad-hoc polymorphism' is understandable: Java supports method, but not operator overloading. Such a challenge is best viewed as a task that can be handled in several ways:

\begin{description}[leftmargin=1em]

\item[Human in the loop] An ontology engineer may dismiss the challenge or refine the ontology to distinguish method from operator overloading.

\item[Auto-close by support] A challenge may be ignored if the fact is strongly supported by other arrows.

\item[Auto-close on repetition] A recurring challenge may be closed automatically if it was previously dismissed.

\item[Arrow tuning] Repeated (arguably vacuous) challenges may motivate prompt refinement to reduce future tasks.

\item[Agentic voting] Different sources or extraction schemes may
  provide richer or alternative semantics to be combined, for example,
  by agentic voting; see Sec.~\ref{sec:bokarrow} for a brief discussion.

\end{description}

\myparagraph{Discussion}

\begin{description}[leftmargin=1em]

\item[ABox only (again)] We considered all language-related properties, but did not assess the TBox itself; this follows in Sec.~\ref{sec:bokarrow}.

\item[Correctness, too] The arrow addresses not only completeness, but also challenges ontology axioms.

\item[Threats to validity] We omitted here explicit knowledge-extraction results, limiting cross-validation.

\end{description}


\subsection{A `\textit{Body of Knowledge} Arrow}
\label{sec:bokarrow}

{\sloppypar

The Software Engineering Body of Knowledge
(\swebok)~\cite{IEEE2024SWEBOK} is an authoritative knowledge
resource on software engineering. Since the domain of interest (\sle)
clearly overlaps with software engineering, it provides a natural
basis for ontology review.

}

\begin{figure}[htbp]
    \vspace{-22\in}
    \centering
    \includegraphics[width=0.98\columnwidth]{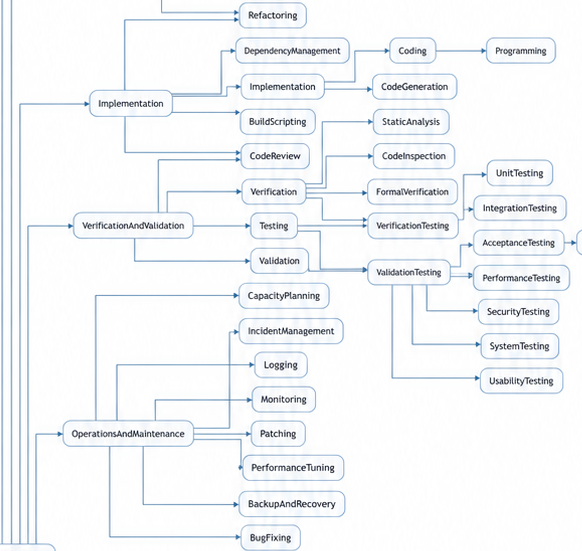}
    \vspace{-37\in}
    \caption{SE activities in the ontology}
    \label{fig:activities}
    \vspace{-37\in}
\end{figure}

\myparagraph{Relevant Ontology Scope}

\fsl{} includes a taxonomy of `software engineering activities',
mainly to associate software languages, tools, formalisms, and related
concepts with specific activities in the software engineering
lifecycle. Figure~\ref{fig:activities} shows an excerpt of this
taxonomy, which contains 94 edges in total.

\myparagraph{Knowledge Extraction}

SWEBOK is primarily a book of several hundred pages; its table of
contents exposes the knowledge areas and their subdivisions relevant
to ontology review. Several derived representations exist at different
levels of formality, governance, and endorsement. We
found one reported effort to extract an ontology from SWEBOK
V3~\cite{ProckoEtAl23} and verified that it captures SWEBOK's knowledge
areas and their decomposition.\footnote{\url{https://github.com/PR0CK0/swebok-rebel-rdf/blob/main/02_single_SWEBOK_file/01-15_SWEBOK_post_construct_FIXED.ttl}}

\begin{figure}[htbp]\scriptsize

\bigskip

\hfill{}Support
\vspace{-99\in}
\begin{lstlisting}[language=json,basicstyle=\ttfamily\scriptsize,numbers=none]
{
"superclass": "AcceptanceTesting",
"subclass": "UserAcceptanceTesting",
"assessment": "support",
"explanation": "SWEBOK represents acceptance/qualification testing as a recognized testing form and repeatedly connects requirements and testing to customers and end users. Treating user acceptance testing as a specialization of acceptance testing is therefore consistent with its decomposition."}
\end{lstlisting}

\medskip

\hfill{}Challenge
\vspace{-99\in}
\begin{lstlisting}[language=json,basicstyle=\ttfamily\scriptsize,numbers=none]
{
"superclass": "EvolutionAndImprovementActivity",
"subclass": "TechnicalDebtManagement",
"assessment": "challenge",
"explanation": "Managing technical debt is compatible with SWEBOK concerns such as maintenance, quality, refactoring, and change, but the resource does not expose technical-debt management as a clear decomposition node of the maintenance knowledge area. The placement is plausible rather than directly confirmed."}
\end{lstlisting}
\vspace{-42\in}
\caption{LLM response for SWEBOK-based ontology review}
\label{fig:swebok-response}
\vspace{-37\in}
\end{figure}

\myparagraph{Ontology review}

Using a prompting model similar to that in
Sec.~\ref{sec:pediaarrow}, we gather supports and challenges. More
than two thirds of the ontology subclassing claims are supported; the
remaining ones are challenged; see Figure~\ref{fig:swebok-response}
for examples. Both explanations are plausible. The challenge may
arise because `technical debt management' is not explicit in SWEBOK
V3. Arguably, it is also not an important activity in the \sle{}
domain. Removing the corresponding activity in response to the
challenge may therefore be sensible; it would reduce noise for
less relevant ontology axioms.

\myparagraph{Discussion}

\begin{description}[leftmargin=1em]

\item[TBox, too] This arrow focuses on the TBox, specifically the
  subclass relationships among SE activities. This form of OWL
  constraint does not introduce additional challenges.

{\sloppypar
  
\item[Multi-agent evaluation] We use a parametrised prompt template
  (such as ``\emph{We aim at a comparison for alignment of two
    resources .... The first resource \{swebok\} is a projection or
    representation of ... SWEBOK .... The second resource \{sle\} is
    a projection or representation of an ontology ...}'').
  This allows alignment to be determined for different projections or
  representations and combined through a voting scheme, thereby
  accounting for intricacies of the resources, especially the size
  and complexity of SWEBOK.

}
  
\item[Threats to validity] The ontology uses a uniform taxonomy for SE
  activities, whereas SWEBOK uses a nuanced decomposition of SE
  knowledge areas; the semantic basis of the alignment is thus
  questionable. Moreover, any useful projection of SWEBOK remains
  considerable in size and may challenge LLM usage.

\end{description}



\newcommand{\os}{\ensuremath{o^{\subseteq}}}
\newcommand{\ot}{\ensuremath{o^{\mathit{Tbox}}}}
\newcommand{\oa}{\ensuremath{o^{\mathit{Abox}}}}
\newcommand{\kt}{\ensuremath{k^{\mathit{Tbox}}}}
\newcommand{\ka}{\ensuremath{k^{\mathit{Abox}}}}
\newcommand{\vt}{\ensuremath{v^{\mathit{Tbox}}}}
\newcommand{\va}{\ensuremath{v^{\mathit{Abox}}}}
\newcommand{\axs}{\ensuremath{\mathcal{A}}}


\section{The \mabbr{} Method}
\label{sec:meth}

In the previous section, we focused on and demonstrated \mabbr{}
arrows, i.e., the `paths' from external resources to affected segments
of an ontology $o$ in domain $d$. An arrow is a review component that
returns assessments supporting or challenging some part of $o$. The
\mabbr{} method administers creation, monitoring, and maintenance of a
system of arrows within a \mabbr{} assurance process.


\subsection{\mabbr{} Arrows}
\label{sec:meth:arrows}

The design and implementation of an arrow $a$ is characterized as
follows:

\begin{enumerate}[leftmargin=1.5em]

\item Identify an \emph{external knowledge resource} $r$ of interest,
  i.e., a resource containing reasonably formal or formalizable
  knowledge in domain $d$. This identification includes an informed
  hypothesis that $r$ and $o$ overlap in a manner useful for ontology
  assurance.

\item Identify a \emph{sub-ontology} \os, thereby projecting $o$ onto
  the scope of $r$ within $o$ and claiming an upper bound on their
  overlap. By default, $\os=o$. The sub-ontology can be derived
  computationally, for example, by a SPARQL query or a DL expression.
  A weak understanding of the scope of $r$ within $o$ limits the
  reliability of challenges produced by $a$.
  
\item Formally represent the relevant knowledge $k$ retrievable from
  $r$ as TBox/ABox segments \kt{} and \ka. At this stage, $o$ and $k$
  need not be aligned; we merely seek a formal representation before
  assessment, thereby enabling additional validation. In practice, we
  may omit this formal representation. For example, the arrow in
  Sec.~\ref{sec:pediaarrow} tasked an LLM with returning assessments
  directly. In such cases, however, we could request a formal
  representation, thereby effectively exercising chain-of-thought
  prompting.

\item Realize arrow $a$ so as to review \os{} relative to $k$ (or $r$).
  Executing $a$ is expected to return a collection of assessments that
  support or challenge \os{} on the grounds of $r$; see below.

\end{enumerate}


\subsection{\mabbr{} Assessments}

An assessment has the following structure:

\begin{itemize}[leftmargin=1em]

\item Type of assessment: support or challenge.

\item The axioms $\axs \subseteq \os$ to which the assessment applies.

\item An optional ontology revision proposal described
  as a state-based (i.e., delta-based) or operation-based
  transformation. In the state-based approach, axioms to be removed
  from and added to $o$ (\os) are described. In the operation-based
  approach, an expression in a suitable ontology transformation
  language is given, whose semantics in turn describes a delta. We
  demonstrate the latter in Sec.~\ref{sec:improve}.

\item An optional confidence value $0 < c \leq 1$ informing, within
  the \mabbr{} process, agents or humans for task management.

\item An optional explanation justifying the assessment, including its
  confidence and revision proposal (if present), while drawing on
  content from $o$ (\os) and $r$ ($k$). Ideally, this explanation
  helps other agents (arrows or humans) with reviewing the assessment.

\end{itemize}

Execution of arrow $a$ can be triggered in different ways:
  
\begin{description}[leftmargin=1em]

\item[Change-triggered] Changes to (1) arrow $a$'s implementation or
  configuration, (2) the knowledge resource $r$ used by the arrow or
  at least its projection to knowledge $k$, and (3) the targeted
  ontology $o$ or at least the ontology scope \os{} trigger
  re-execution of $a$.

\item[User-triggered] The process owner (an ontology engineer)
  explicitly initiates execution; this may be useful during
  development and maintenance of the arrow.

\item[Time-triggered] The arrow is (re-)executed regularly (e.g.,
  during a nightly build); this is useful when change tracking, explicit
  knowledge extraction (going from $k$ to $r$), or ontology scoping
  (going from $o$ to \os) is limited.

\end{description}


\begin{table}[htbp]
\centering\footnotesize
\begin{tabular}{|p{0.46\columnwidth}|p{0.16\columnwidth}|p{0.24\columnwidth}|}
  \hline
\textbf{External-resource situation} &
\textbf{Assessment} &
\textbf{Typical revision proposal} \\
\hline
  \hline
 Resource corroborates ontology content &
support &
none \\
  \hline

Resource provides contrary evidence &
challenge &
correction \\
  \hline

Resource provides positive evidence for content missing from the ontology &
challenge &
completion \\
  \hline

Resource provides evidence of inadequate conceptual organization &
challenge &
restructuring \\
  \hline

Resource is silent, and coverage is unknown or incomplete &
uncovered &
none \\
  \hline

Resource is silent, but coverage is assumed complete for this axiom type &
challenge &
correction \\
\hline
\end{tabular}
\caption{Assessment options for arrows}
\label{tab:assessment-options}
\vspace{-37\in}
\end{table}


Revision proposals may call for correction, completion, or
restructuring, the latter including refactoring and
reconceptualization. Table~\ref{tab:assessment-options} characterizes the
corresponding assessment options.

{\sloppypar
  
Importantly, we take an essentially model-based,
software engineering-like view of ontology revision. In particular,
determining ontology consistency (as an internal logical property) and
adequacy (in terms of competency questions) is not the focus.
Instead, \mabbr{} ontology reviews aim to support or challenge axioms,
where challenges may call for completion (adding axioms), correction
(removing axioms), or restructuring (rewriting sets of axioms). We
also make no significant distinction between TBox and ABox here. For
example, subclass relationships are `just' axioms that can be
supported or challenged like ordinary ABox assertions.

}

At the process level, \mabbr{} assessments essentially become `tasks'
in the sense of task management.


\subsection{Quantified Assurance}

The \mabbr{} method relies on creating, monitoring, and maintaining a
system of arrows within a \mabbr{} assurance process. We quantify
assurance similarly to coverage in software testing: namely, by
measuring to what extent and in what way the ontology in a given
process state is covered by arrows. The basic coverage situations at
the per-axiom level are as follows:

\begin{description}[leftmargin=1em]

\item[Support-only coverage] The axiom is supported by at least one
  arrow and has no pending challenges; if such tasks existed, they
  were closed as `Won’t Fix'.

\item[Net-positive coverage] The axiom is more supported than
  challenged in the current process state; we discuss aggregation of
  assessments below.

\item[Net-negative coverage] The axiom is more challenged than
  supported in the current process state.

\item[Challenge-only coverage] The axiom is challenged by at least one
  arrow and has no rivaling support. The challenges must be addressed,
  and corresponding revision proposals must be considered.

\item[Missing coverage] The axiom is neither supported nor challenged.
  There is no assurance for this element. The system of arrows should
  be advanced to achieve coverage.

\end{description}

Challenges and supports are aggregated via the Abelian group
$(\mathbb{R}, +, -, 0)$ using their associated confidences, with
positive reals for supports and negative reals for challenges. For
example, a support with confidence $0.75$ stacked against a challenge
with confidence $0.50$ yields net-positive assurance of $0.25$. Other
schemes for aggregating per-axiom assessments are not discussed here
for brevity.

We speak of \emph{complete assurance} if all ontology axioms achieve
\emph{support-only coverage} or \emph{net-positive coverage}.
Otherwise, we speak of \emph{incomplete assurance}. We quantify
assurance by a 5-tuple
\[\langle \mathit{soc}, \mathit{npc}, \mathit{nnc}, \mathit{coc},
  \mathit{mc} \rangle,\]
with the number of ontology axioms $\mathit{soc}$
having support-only coverage, $\mathit{npc}$ net-positive coverage,
$\mathit{nnc}$ net-negative coverage, $\mathit{coc}$ challenge-only
coverage, and $\mathit{mc}$ missing coverage. Other ontology-level
assurance metrics are not discussed here for brevity.

Using a human in the loop and agentic techniques, we can close tasks
arising from challenges; these tasks can also be auto-closed if they
reappear in subsequent executions. In this manner, we can often move
from net-positive to support-only coverage. Alternatively, the arrows
causing the challenges may require refinement rather than manual or
agentic override.


\subsection{Advanced \mabbr{} Issues}

Several issues make the \mabbr{} process challenging to implement and
execute efficiently, particularly given the need to minimize work for
the human in the loop and avoid redundant effort. We mention several
issues here and defer a more comprehensive discussion to an extended
presentation.

\begin{description}[leftmargin=1em]

\item[Cross validation] Claiming full assurance based on a single
  supportive arrow may be optimistic, depending on the axiom type, the
  quality of the knowledge resource, and the arrow implementation,
  none of which is reliably captured by confidence values. We can
  therefore require cross validation (multiple supports) in addition
  to full assurance.

\item[Confidence calibration] Confidence values are decisive when
  distinguishing net-positive from net-negative coverage. Ontology
  engineers must be particularly alert to overconfidence in supports
  and underconfidence in challenges.

\item[Explainable assurance] Assurance can be quantified by the
  5-tuple defined above, but an ontology engineer must also understand
  coverage qualitatively, for example, when seeking new scopes for
  coverage through revised or additional arrows. Such `explainable
  assurance' can rely specifically on the visualization of coverage.

\end{description}



\section{Demonstration of Assurance Improvement}
\label{sec:improve}

We demonstrate assurance improvement for \fsl{} with respect to a
specific, easy-to-understand aspect~---~entity linkage to Wikipedia.
We present the improvement end to end, starting from an initial
(suboptimal) assurance situation, followed by LLM-based linkage
discovery with auxiliary steps for context extraction, prompt
generation, and execution. This leads to task generation in the
\mabbr{} sense and semi-automatic task resolution, after which we
assess the final, improved assurance situation.


\subsection{Initial Assurance}

Our `Wikipedia Linkage Policy' validator, at some point back, reported the following
numbers:

\begin{itemize}
\item There are \emph{600} entities (languages, tools, formalisms, etc.) under the policy.
\item \emph{16} entities are `opt-out' marked (link N/A).
\item \emph{113} entities lack any linkage.
\item \emph{86} entities have invalid linkage.
\item \emph{405} entities (67.5\,\%) have valid linkage.
\item foaf:isPrimaryTopicOf versus foaf:page usage is 165/324.
\end{itemize}

Here, valid linkage means that English Wikipedia articles are used,
with exactly one article per resource; validity does not establish the
suitability of the link. The numbers clearly leave room for assurance
improvement: one third of the entities are not validly linked.
Arguably, there is also excessive use of the semantically weaker
`foaf:page' predicate instead of the stronger
`foaf:isPrimaryTopicOf'.


\subsection{Context Extraction}

We seek an LLM-based arrow that assesses Wikipedia linkage, both to
validate the suitability of existing links and to propose suitable
links where they are missing. We assume that the LLM must observe the
resource to be linked and the ontology to make a reasonable linkage
judgment.

\begin{figure}[htbp]\scriptsize
\begin{lstlisting}[language=Sparql,basicstyle=\ttfamily\scriptsize,numbers=none]
SELECT DISTINCT ?entity ?direction ?subject ?predicate ?object
WHERE {
  ?entity rdf:type/rdfs:subClassOf* tbox:Entity .
  FILTER(isIRI(?entity))
  {
    ?entity ?predicate ?object .
    BIND(?entity AS ?subject)
    BIND("outgoing" AS ?direction)
  }  UNION {
    ?subject ?predicate ?entity .
    BIND(?entity AS ?object)
    BIND("incoming" AS ?direction)
  }
}
\end{lstlisting}
\vspace{-42\in}
\caption{Sparql query for context extraction.}
\label{fig:context}
\vspace{-37\in}
\end{figure}

Resources of interest derive from `tbox:Entity'. For each such entity,
we collect all triples in which it occurs as subject or object,
classifying them accordingly as outgoing or incoming.
Figure~\ref{fig:context} shows the essential query used for each
entity. The result is captured in JSON for use in
prompting.{}
This context extraction constitutes a particular approach to
sub-ontology identification according to Sec.~\ref{sec:meth:arrows}.


\subsection{LLM-based Assessment}

Here is the prompt template (compressed for presentation) used for
each entity:

\begin{quote}\itshape\footnotesize
Assess Wikipedia linkage for an ontology resource using its
attached knowledge-graph context, the ontology TBox, and Wikipedia
search. Identify existing foaf:isPrimaryTopicOf or foaf:page links and
return exactly one policy-compliant canonical English Wikipedia link,
preferring foaf:isPrimaryTopicOf when the article is primarily about
the resource and otherwise using foaf:page; use the designated N/A
fallback if no suitable article exists. Classify the result as support
when the context contains exactly one compliant link, replace when
existing links must be consolidated or changed, or resolve when no
link exists. Output only the prescribed, directly parseable JSON
object containing the resource, assessment, links found, and—except
for support—the proposed link.
\end{quote}

\begin{table}[htbp]
\vspace{-37\in}
\centering\footnotesize
\begin{tabular}{|l|r|} \hline
  \textbf{Assessment} & \textbf{Frequency}\\\hline
  support & 287\\
  replace & 204\\
  resolve & 109 \\\hline
\end{tabular}
\caption{Assessment summary for Wikipedia linkage.}
\label{tab:assessment-summary}
\vspace{-37\in}
\end{table}

Table~\ref{tab:assessment-summary} summarizes the assessments resulting
from applying the template to all entities in
scope.


\subsection{Task Processing}

Several aspects of the LLM-based assessments and accompanying ontology
revision proposals are algorithmically checkable and implemented as
companion components of the Wikipedia-linkage arrow: checks for the
existence of articles for proposed links, consistency between links
observed by the LLM and those actually present in the ontology, and
consistency between the number of links found and the assessment type
(e.g., `resolve' must correspond to 0 links present). We rely on a
human in the loop for further validation as follows.

\begin{itemize}
  
\item `Support' assessments are not further validated, as we assume
  that links present in the ontology were deliberately injected and
  are now further backed up by an `expert' (an LLM).

\item `Replace' assessments are reviewed by the human in the loop. In
  this instance, almost all replacements arise from multiple links
  being present and/or the use of DBpedia instead of
  Wikipedia. All replacements were confirmed by the human in the loop.

\item `Resolve' assessments are reviewed for relevance and for the
  suitability of isPrimaryTopic versus page. We reviewed the first 10
  assessments and then confirmed the entire package, delegating more
  in-depth relevance/suitability validation to another \mabbr{} arrow,
  not discussed in this short paper.
    
\end{itemize}

In total, \emph{1117} automatically synthesized Sparql queries/updates
were performed, resulting in an ontology with improved Wikipedia
linkage; see below.


\subsection{Final Assurance}

Our `Wikipedia Linkage Policy' validator reports the following
numbers on the evolved ontology:

\begin{itemize}
\item All \emph{600} entities under the policy have valid linkage.
\item \emph{17} entities are `opt-out' marked (link N/A).
\item \emph{0} entities lack any linkage.
\item \emph{0} entities have invalid linkage.
\item foaf:isPrimaryTopicOf/page usage is 244/356.
\end{itemize}

Thus, LLM-based linkage assessment, human-in-the-loop proposal
approval, and Sparql-based ontology modification result in a fully
satisfactory ontology with respect to the Wikipedia Linkage Policy.

\section{\sle{} Knowledge Resources}
\label{sec:cata}

We identify external knowledge resources for systematic review and
maintenance of an \sle{} domain ontology. We seek resource
\emph{categories} and concrete \emph{representatives} providing
evidence about terminology, conceptual organisation, constraints,
individuals, and relations.

We distinguish seven resource categories plus supporting tools.
Categories may overlap; placement reflects their primary review
contribution. The demonstrations from Sec.~\ref{sec:arrows} are
included among the representatives. For brevity, we give one
representative per category;
\ifanonymous{}see the reproducibility package\fi
{} for additional representatives.


\subsection{Technology Inventories and Ecosystem Metadata}
\label{sec:cata:inventories}

These resources describe concrete technologies such as languages,
tools, implementations, and packages, including registries,
popularity indices, package catalogues, and services tracking
dependencies, releases, compatibility, and support.

{\sloppypar
  
\paragraph{Representative:} TIOBE Index\footnote{\url{https://www.tiobe.com/tiobe-index/}}:
programming languages with monthly popularity indicators.

}

\begin{description}[leftmargin=1em]
  \item[Review contribution:] Identify candidate language instances
    and prioritise coverage checks.
  \item[Preparation:] Extract language lists and reconcile names
    with ontology identifiers.
  \item[Qualification:] Popularity informs completeness;
    nonpopularity does not imply incorrectness (irrelevance).
\end{description}


\subsection{Encyclopaedias and Reference Works}
\label{sec:cata:reference}

These resources describe software languages, features, and related
concepts through explanatory text and structured fields, providing
evidence for linking concrete languages to ontology concepts.

{\sloppypar

\paragraph{Representative:} Wikipedia articles on individual
    programming languages\footnote{\url{https://en.wikipedia.org/wiki/Lists_of_programming_languages}},
    including descriptions of language features and paradigms.

 }
  
\begin{description}[leftmargin=1em]
  \item[Review contribution:] Identify concepts evidenced by
    language descriptions and review their association with
    the corresponding ontology language.
  \item[Preparation:] Match articles to language entities,
    extract relevant statements, and map terminology to ontology
    concepts.
  \item[Qualification:] Concept associations require contextual
    interpretation; an unmentioned feature cannot be assumed absent.
\end{description}


\subsection{Standards and Specifications}
\label{sec:cata:standards}

These resources define languages, formats, interfaces, and conformance
requirements, providing version-specific evidence about constructs,
semantics, and constraints.

{\sloppypar

\paragraph{Representative:} ECMA-262~\cite{cata:ecma262}, the ECMAScript
    language
    specification\footnote{\url{https://ecma-international.org/publications-and-standards/standards/ecma-262/}}.

}
    
\begin{description}[leftmargin=1em]
  \item[Review contribution:] Check language terminology, constructs,
    and specified properties.
  \item[Preparation:] Select an edition and map specification terms
    and rules to ontology elements.
  \item[Qualification:] Normative definitions must be distinguished
    from implementation behaviour and extensions.
\end{description}


\subsection{Bodies of Knowledge}
\label{sec:cata:bok}

These resources synthesise a discipline's accepted knowledge into
areas and topics, supporting review of disciplinary scope and the
organisation of engineering activities and methods.

{\sloppypar

\paragraph{Representative:} SWEBOK~\cite{cata:swebok3}\footnote{\url{https://www.computer.org/education/bodies-of-knowledge/software-engineering}}:
software engineering knowledge areas and their decomposition.

}

\begin{description}[leftmargin=1em]
  \item[Review contribution:] Review coverage and organisation of
    SE areas and activities.
  \item[Preparation:] Match areas and topics to ontology concepts;
    retain provenance of derived RDF representations.
  \item[Qualification:] Topic decomposition does not itself establish
    activity subclassing; derived representations require source checking.
\end{description}


\subsection{Taxonomies and Controlled Vocabularies}
\label{sec:cata:vocabularies}

Classifications of language paradigms, features, processors, and
formal foundations provide terms and relationships for reviewing SLE
concepts; broad classifications are considered through selected
branches.

{\sloppypar

\paragraph{Representative:} The language-related branches of ACM
    CCS~\cite{cata:acmccs}\footnote{\url{https://www.acm.org/publications/class-2012}},
    particularly general programming languages and compilers
    under ``Software notations and tools''.

}
  
\begin{description}[leftmargin=1em]
  \item[Review contribution:] Compare coverage and organisation
    of language types, features, and implementation concepts.
  \item[Preparation:] Delimit selected branches, match terms to
    ontology concepts, and interpret their relationships.
  \item[Qualification:] Subject-indexing hierarchies do not
    automatically express subclassing; selected branches provide
    only partial SLE coverage.
\end{description}


\subsection{Research Literature and Bibliographic Resources}
\label{sec:cata:literature}

SLE publications provide evidence about language design,
implementation, evolution, methods, and tools. Venue-defined
collections offer focused starting points, complemented by
topic-based searches.

{\sloppypar

\paragraph{Representative:} Proceedings of the International Conference
    on Software Language Engineering (SLE)\footnote{\url{https://www.sleconf.org/}},
    indexed by
    dblp\footnote{\url{https://dblp.org/db/conf/sle/index.html}}.

}
  
\begin{description}[leftmargin=1em]
  \item[Review contribution:] Identify language-engineering
    concepts, methods, and tools, and publications substantiating
    their properties and relationships.
  \item[Preparation:] Select proceedings years, screen papers
    for relevance, and extract full-text evidence.
  \item[Qualification:] Titles support discovery, not substantive
    claims; a single venue does not exhaust SLE research.
\end{description}
  

\subsection{Software Artifacts and Empirical Collections}
\label{sec:cata:empirical}

Implementations, examples, test suites, benchmarks, corpora, and
datasets provide concrete evidence of language features and
engineering practices.

{\sloppypar

\paragraph{Representative:} Computer Language Benchmarks
    Game\footnote{\url{https://benchmarksgame-team.pages.debian.net/benchmarksgame/}}:
    benchmark tasks, programs, and measurements.

}

\begin{description}[leftmargin=1em]
  \item[Review contribution:] Identify languages and implementations
    and inspect concrete examples of their use.
  \item[Preparation:] Identify languages, implementations, and
    features evidenced by selected programs; map them to ontology
    entities and concepts for comparison with existing assertions or
    candidate additions.
  \item[Qualification:] A program may witness a feature's use, but
    attribution to the language requires distinguishing language
    constructs from library facilities and implementation extensions.
    Absence from the collection does not establish lack of support.
\end{description}


\subsection{Evidence Processing and Interlinking Tools}
\label{sec:cata:tools}

These tools support extraction, verification, aggregation,
normalisation, and reconciliation while keeping evidence attributable
to its underlying sources.

{\sloppypar

\paragraph{Representative:} OpenRefine\footnote{\url{https://openrefine.org/}}:
data cleaning, transformation, and reconciliation.

}

\begin{description}[leftmargin=1em]
  \item[Review contribution:] Prepare resource records and match
    entities across resources and the ontology.
  \item[Preparation:] Import records and configure transformations
    and a suitable reconciliation service.
  \item[Qualification:] Candidate matches require validation;
    reconciliation does not independently substantiate domain claims.
\end{description}

\section{Related Work}
\label{sec:related}

\myparagraph{Ontology and knowledge-graph evaluation}

Ontology evaluation encompasses structural, functional, and usability
criteria~\cite{gangemi2006modelling}, as well as comparison with reference
ontologies, corpora, tasks, and explicit criteria~\cite{raad2015survey}.
OOPS! detects recurrent modelling pitfalls in ontologies~\cite{poveda2014oops}.  RDFUnit derives executable quality tests from
schemas and test patterns, tracks their coverage, and supports repeated
evaluation of Linked Data~\cite{kontokostas2014test}; SHACL checks RDF
graphs against declared constraints~\cite{w3c2017shacl}.  These approaches
are relevant to the executable and coverage-oriented aspects of \mabbr.
\mabbr's proposed assessment differs in its reference point: an arrow
compares a selected part of a maintained ontology with a particular external
domain resource, whose scope and authority must be judged before a support
or challenge is interpreted.  Consequently, a failed constraint, absence
from a resource, and contrary domain evidence should not be treated as
equivalent findings.

\vspace{-22\in}

\myparagraph{External evidence and ontology change} Ontology debugging
and completion distinguish defect detection from repair, including the
problem of choosing appropriate repairs~\cite{lambrix2023completing};
knowledge-graph refinement also considers error detection and
completion~\cite{paulheim2017knowledge}.  Especially close to \mabbr,
Evolva discovers candidate knowledge from heterogeneous external
sources and uses background knowledge to support ontology adaptation
and the management of changes~\cite{zablith2009evolva}.
Process-oriented accounts of ontology evolution likewise include
detecting changes and validating their
implementation~\cite{zablith2015ontology}.  \mabbr{} builds on this
line of research by organising systems of knowledge extraction and
ontology review components in a process; a \emph{review} can support
an existing axiom, challenge one, or motivate completion or
restructuring, without committing every extracted statement to the
ontology. 

\vspace{-22\in}

\myparagraph{Reconciliation and AI-assisted review} Correspondences
between resource entities and ontology entities are a prerequisite for
many reviews: ontology matching studies such
correspondences~\cite{euzenat2013ontology}, and link-discovery systems
such as Silk find and maintain links between data
sources~\cite{volz2009silk}.  These methods can help implement an
arrow, but a match alone does not verify a domain assertion or
establish that a Wikipedia article is an appropriate primary topic for
an ontology entity.  Recent LLM-assisted ontology evaluation includes
OE-Assist, which investigates automated and human-assisted
verification of competency questions~\cite{lippolis2025assisting}.
\mabbr{} instead uses resource-grounded assessments of ontology
content, including TBox and ABox assertions; its Wikipedia and SWEBOK
arrows illustrate this design.  LLM-produced explanations and
confidence values remain candidate evidence for review, not
independently verified guarantees of correctness.

\vspace{-22\in}

\myparagraph{Workflow and traceability}
OnToology automates ontology engineering support activities in repository
workflows~\cite{alobaid2019ontoology}, while ROBOT supports automated
ontology workflows~\cite{jackson2019robot}.  PROV-O supplies terms for
recording entities, activities, agents, and their relationships~\cite{lebo2013provo}.  Such tools and representations can support the
execution and traceability of \mabbr{} arrows.  The proposed contribution is
the organization of heterogeneous domain resources into recurring,
scoped reviews whose assessments can be linked to ontology elements and
considered as revision tasks.  The present case study illustrates this
process; the reliability of assessments and the effort required for human
triage remain questions for further evaluation.


\section{Conclusion}
\label{sec:concl}

We have presented \mabbr{}, a method for continuous ontology assurance
through external resources. Executable, resource-scoped arrows support
or challenge ABox and TBox axioms, generate revision tasks, and enable
per-axiom assurance coverage. In the case study based on the \fsl{}
ontology, arrows review, for example, language coverage, language
assertions, software-engineering activities, and linkage policy. We
illustrated \mabbr-based improvement of assurance in an LLM-assisted,
human-triaged Wikipedia-linkage exercise which raised compliance with
the defined policy from 405 to 600 of 600 entities. We presented a
catalogue of \sle{} resources describing the review landscape
comprehensively.

Future work is concerned with advanced process tooling, visualisation
of coverage (assurance), cross-validation of arrows, and measurement
or optimization of human triage effort over time.


\vspace{-22\in}

\section*{Acknowledgment}

{\small
  
\noindent
Generative AI tools were used for editorial assistance with the manuscript and to support literature discovery alongside searches of scholarly databases. They were not used in the study design, data collection, analysis, interpretation, or any other methodological aspect of the underlying research. The author(s) take(s) full responsibility for the content of this paper.

}

\bibliographystyle{ACM-Reference-Format}
\bibliography{bibs/paper,bibs/cata,bibs/appcata,bibs/related}

\end{document}